\documentclass[11pt]{article}

\usepackage[final]{acl}

\usepackage{times}
\usepackage{latexsym}
\usepackage[T1]{fontenc}
\usepackage[utf8]{inputenc}
\usepackage{microtype}
\usepackage{inconsolata}

\usepackage{graphicx}
\usepackage{booktabs}
\usepackage{array}
\usepackage{amsmath}
\usepackage{amssymb}
\usepackage{mathtools}
\usepackage{float}
\usepackage{caption}
\usepackage{needspace}
\usepackage[capitalize,noabbrev]{cleveref}

\hypersetup{
  pdftitle={Arabic Safety Alignment as Selective Refusal: An Empirical Study of SFT, DPO, and Guard Calibration},
  pdfauthor={Mohamad Zbib, Mohamad Bazzi, and Ammar Mohanna}
}

\newcommand{\safe}{B}
\newcommand{\unsafe}{H}

\title{Arabic Safety Alignment as Selective Refusal:\\ An Empirical Study of SFT, DPO, and Guard Calibration}

\author{
  Mohamad Zbib\thanks{\ \ Equal contribution.} \\
  American University of Beirut \\
  \texttt{mbz02@mail.aub.edu}
  \And
  \And
  Mohamad Bazzi\footnotemark[1] \\
  American University of Beirut \\
  \texttt{mab101@aub.edu.lb}
  \And
  \And
  Ammar Mohanna \\
  American University of Beirut \\
  \texttt{am288@aub.edu.lb}
}

\begin{document}
\maketitle

\begin{abstract}
Arabic large language models must refuse harmful prompts without over-refusing benign or sensitive prompts, yet a single refusal rate hides this trade-off. We evaluate it using benign refusal $\safe$ and harmful-prompt refusal $\unsafe$, where $\unsafe$ measures refusal rather than harmful compliance. Across five Arabic-capable models and 130 runs on the full human-written AraSafe set, refusal-only supervised fine-tuning (SFT) collapses toward blanket refusal, whereas selected mixed-SFT configurations reach $\unsafe\approx90$--93\% at $\safe=14$--23\%; four selected configurations exceed $\unsafe=90\%$ in all three runs, while Fanar does so in two of three. Direct Preference Optimization (DPO) and inference guards change $\safe$ and $\unsafe$ differently across models rather than acting as uniform upgrades. In a blinded 300-response audit, annotator binary-refusal agreement is 89.0\% ($\kappa=0.78$); Qwen3Guard and Aya Expanse 32B reach 88.7\% and 91.0\% accuracy, respectively, with no conclusive paired difference. Selected SFT raises $\unsafe$ on Arabizi for all five models, but none reaches 90\%, showing only partial transfer from Modern Standard Arabic; overall, the results support model-specific operating-point selection: set a deployment target and retain only interventions that improve it.
\end{abstract}

\section{Introduction}

Arabic safety is not a single input space. Users write in Modern Standard Arabic (MSA), regional dialects such as Egyptian and Levantine, romanized Arabizi, and noisy text, and many benign prompts are sensitive: news, medical, legal, and policy questions that share vocabulary with genuinely harmful requests \citep{ashraf-etal-2025-arabic, mousi2024aradice}. A safety method for Arabic must therefore refuse harmful prompts while staying usable across these forms. A single refusal rate cannot capture this requirement because a model can appear protective simply by refusing everything. The practical question is whether standard alignment interventions improve this trade-off consistently across Arabic-capable models and writing forms.

We evaluate selective refusal with two rates: benign refusal $\safe=P(\mathrm{refusal}\mid\mathrm{benign\ prompt})$ and harmful-prompt refusal $\unsafe=P(\mathrm{refusal}\mid\mathrm{harmful\ prompt})$, measured on the human-written AraSafe benchmark \citep{mubarak2025arasafe}. Crucially, $\unsafe$ records whether a response refuses; its complement is not automatically harmful compliance. A useful intervention must lower $\safe$ or raise $\unsafe$ without silently damaging the other axis, so every method is judged by movement on the $(\safe,\unsafe)$ plane rather than either rate alone \citep{rottger-etal-2024-xstest, cui2024orbench}.

\begin{figure}[t]
  \centering
  \includegraphics[width=\linewidth]{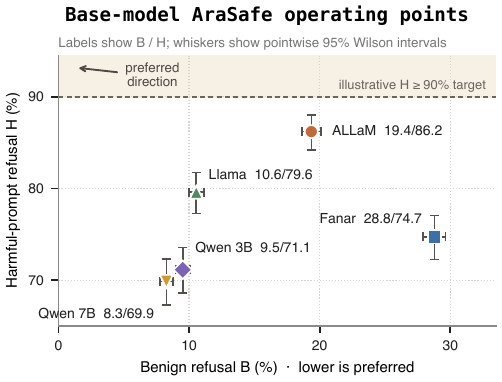}
  \caption{Base-model operating points on AraSafe. Lower benign refusal $\safe$ and higher harmful-prompt refusal $\unsafe$ are preferred; labels report $\safe/\unsafe$. Whiskers show marginal 95\% Wilson intervals for prompt sampling conditional on the automatic judge labels ($n_B=10{,}823$, $n_H=1{,}254$), reconstructed from retained rates. The dashed line marks the illustrative $\unsafe\ge90\%$ target, not a universal safety standard.}
  \label{fig:scatter}
\end{figure}

\Cref{fig:scatter} motivates the study: five Arabic-capable models begin at different measured locations in the $(\safe,\unsafe)$ plane. This variation motivates model-by-model testing; it does not establish universal model categories or transferable intervention rules. Our central claim is narrower: \emph{Arabic selective refusal is a model-specific operating-point selection problem: measure benign refusal $\safe$ and harmful-prompt refusal $\unsafe$, specify the deployment target, and retain an intervention only when it improves that target.}

\paragraph{Contributions.}
\begin{itemize}\setlength{\itemsep}{1pt}
  \item \textbf{Audited evaluation.} We audit $\safe/\unsafe$ labels with 300 blinded responses and a cross-family judge.
  \item \textbf{Model-specific SFT trade-offs.} Refusal-only SFT collapses, while selected mixed-SFT configurations reach $\unsafe\approx90$--93\% at $\safe=14$--23\%.
  \item \textbf{No uniform post-training upgrade.} DPO and guards move models differently; ordering is a null result.
  \item \textbf{Partial cross-script transfer.} Selected SFT raises Arabizi $\unsafe$ for every model, but none reaches 90\%.
\end{itemize}

\section{Measuring and Validating Selective Refusal}
\label{sec:protocol}

\paragraph{Metrics and operating points.} An operating point is one measured pair $(\safe,\unsafe)$. One tested point dominates another when it has lower or equal $\safe$ and higher or equal $\unsafe$, with at least one strict inequality; the tested frontier is the non-dominated subset of observed points. We use $\unsafe\ge90\%$ as an illustrative selection constraint and choose the lowest-$\safe$ feasible candidate. This is not a universal safety standard: a deployment owner must set the target from its own costs, and we report sensitivity to 85\%, 90\%, and 95\% in \cref{tab:threshold-sensitivity}.

\paragraph{Evaluation set and judge.} We evaluate on the full human-written portion of AraSafe \citep{mubarak2025arasafe}: 12,077 prompts, of which 10,823 are benign and 1,254 harmful. We reserve these prompts for evaluation; none appears in training, direction extraction, or calibration. Because AraSafe is MSA-heavy, we treat the operating points in \cref{sec:sft,sec:calibration} as MSA measurements and stress-test them across forms in \cref{sec:robustness}. Qwen3Guard-Gen-4B \citep{zhao2025qwen3guard, souly2024strongreject} supplies the binary refusal label.

\paragraph{Human and cross-judge validation.} We audited a blinded, stratified sample of 300 responses with two independent annotators and adjudication. Binary-refusal agreement was 89.0\% ($\kappa=0.78$). Against adjudicated labels, Qwen3Guard achieved 88.7\% accuracy and 0.887 macro-F1, with a clustered-bootstrap 95\% accuracy interval of [85.3, 91.7]; Aya Expanse 32B \citep{dang2024ayaexpanse} achieved 91.0\% accuracy and 0.910 macro-F1. Their paired difference was inconclusive (McNemar $p=0.382$; Qwen-minus-Aya macro-F1 difference $-2.3$ points, 95\% interval [$-6.4$, 2.0]). Qwen3Guard accuracy was 86.7\% on Qwen-family outputs and 90.0\% otherwise; the gap interval [$-3.3$, 10.3] neither establishes nor excludes family bias. Reported form-specific accuracies range from 87.5\% to 95.0\% on MSA, Egyptian, Levantine, and noisy Arabic subsets, but fall to 77.5\% on Arabizi, where macro-F1 is 0.763. Full audit statistics appear in \cref{tab:judge-audit}.

\paragraph{Models.} The five base models span sizes and providers: Qwen2.5-3B and Qwen2.5-7B Instruct, Meta-Llama-3-8B Instruct, Fanar-1-9B, and ALLaM-7B Instruct. Two are Arabic-centric and three are multilingual; neither provenance group occupies one common starting location in \cref{fig:scatter}. Full identifiers are in Appendix~\ref{app:ids}.

\paragraph{Training and preference data.} Harmful SFT and preference prompts come from an MSA translation of BeaverTails \citep{ji2023beavertails}, translated with the Hala English--Arabic models \citep{hammoud2025hala}; benign SFT examples are separately drawn from Hala-4.6M-SFT. DPO uses 10K benign and 10K harmful preference pairs: harmful pairs rank a refusal over a harmful completion, while benign pairs rank either two helpful answers (V1) or a helpful answer over a refusal (V2) from the Arabic Data Is Better Together collection \citep{dibt2024ar}. Because the harmful supervision is translated MSA, cross-script transfer is an empirical question rather than an assumption.

\paragraph{Uncertainty and seeds.} Wilson intervals quantify prompt-sampling uncertainty conditional on the automatic labels; they do not include judge measurement error, which the audit exposes separately. With 10,823 benign and 1,254 harmful prompts, $\safe$ half-widths are below one point at the reported rates; $\unsafe$ half-widths are about 1.5 points near the selected $\unsafe\approx90\%$ points and 1.8--2.6 points at the base points in \cref{fig:scatter}. Main-text rates are rounded to one decimal place, while appendix tables retain the available source precision; a rounded boundary value is not treated as stable evidence of threshold satisfaction. Primary sweeps use seed 42. Selected mixed-SFT configurations were repeated with seeds 43 and 44: across models, standard deviations span 0.13--0.57 points for $\safe$ and 0.41--1.10 for $\unsafe$. ALLaM, Llama, Qwen 3B, and Qwen 7B exceed $\unsafe=90\%$ in all three runs; Fanar does so in two of three, with mean $\unsafe=89.95\pm0.41$, and is therefore threshold-sensitive. Direction AUC intervals use the Hanley--McNeil estimator.

\paragraph{Scope of the starting points.} The five observed base points differ substantially: ALLaM starts with comparatively high $\unsafe$, Fanar with the highest $\safe$, and the Qwen and Llama models with lower $\safe$ but also lower $\unsafe$. These are measured operating points, not discovered clusters, causal diagnoses, or intervention prescriptions. We therefore search candidate operating points separately for each model. \Cref{tab:notation} makes the paper's notation and inferential scope explicit, while \cref{tab:data-roles} records which data touch each stage.

\begin{table}[t]
  \caption{Notation and scope. Refusal is measured independently on benign and harmful prompts; $\unsafe$ is not a harmful-compliance rate.}
  \label{tab:notation}
  \centering
  \footnotesize
  \setlength{\tabcolsep}{2pt}
  \begin{tabular}{@{}>{\raggedright\arraybackslash}p{0.22\linewidth}>{\raggedright\arraybackslash}p{0.68\linewidth}@{}}
    \toprule
    Term & Meaning \\
    \midrule
    $\safe$ & $P(\mathrm{refusal}\mid\mathrm{benign\ prompt})$; false-positive refusal rate \\
    $\unsafe$ & $P(\mathrm{refusal}\mid\mathrm{harmful\ prompt})$; refusal, not harmful compliance \\
    Operating point & One measured pair $(\safe,\unsafe)$ for a model and intervention \\
    Tested frontier & Non-dominated subset of the operating points actually evaluated \\
    \bottomrule
  \end{tabular}
\end{table}

\begin{table}[t]
  \caption{Data roles in the operating-point search. Evaluation prompts are held out from training and calibration; the remaining stages use the sources and splits listed below.}
  \label{tab:data-roles}
  \centering
  \scriptsize
  \setlength{\tabcolsep}{2pt}
  \begin{tabular}{@{}p{0.27\linewidth}p{0.24\linewidth}p{0.39\linewidth}@{}}
    \toprule
    Split & Used for & Why it matters \\
    \midrule
    AraSafe & Main $\safe/\unsafe$ evaluation & Measures Arabic selective refusal \\
    SFT and DPO data & Training & Supplies helpful and refusal pressure \\
    Boundary set & Stress test only & Checks dialect, Arabizi, noise, and sensitive benign generalization \\
    Activation split & Direction extraction & Learns candidate linear refusal directions \\
    Calibration split & Threshold tuning & Selects per-model thresholds separately from extraction \\
    \bottomrule
  \end{tabular}
\end{table}

\section{SFT Candidate Search}
\label{sec:sft}

\begin{figure*}[t]
  \centering
  \includegraphics[width=\linewidth]{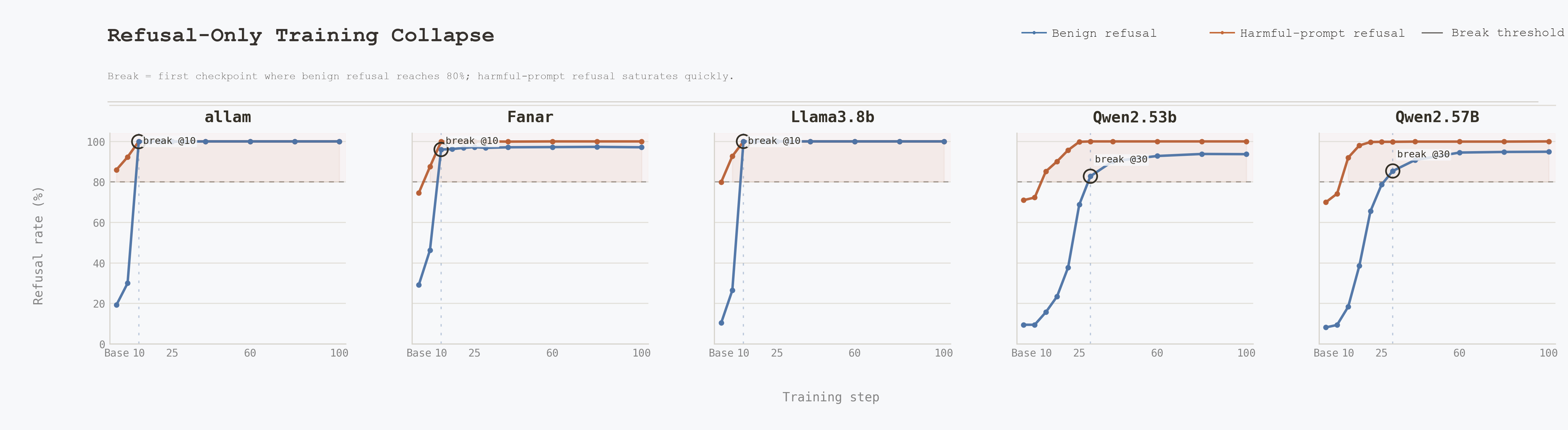}
  \caption{Refusal-only SFT collapses toward blanket refusal. In each model panel, benign refusal $\safe$ (blue) and harmful-prompt refusal $\unsafe$ (orange) are plotted over training steps. The dashed horizontal line marks $\safe=80\%$; the circled marker identifies the first crossing.}
  \label{fig:collapse}
\end{figure*}

\paragraph{Refusal-only SFT collapses toward blanket refusal.} We use refusal-only training \citep{qi2024finetuning} as a diagnostic. As \cref{fig:collapse} shows, harmful-prompt refusal $\unsafe$ saturates within a few steps, while benign refusal $\safe$ rises toward total refusal: ALLaM, Fanar, and Llama cross $\safe=80\%$ by step 10, and both Qwen models by step 30. Thus, refusal pressure must be bounded by benign behavior.

\begin{figure*}[t]
  \centering
  \includegraphics[width=\linewidth]{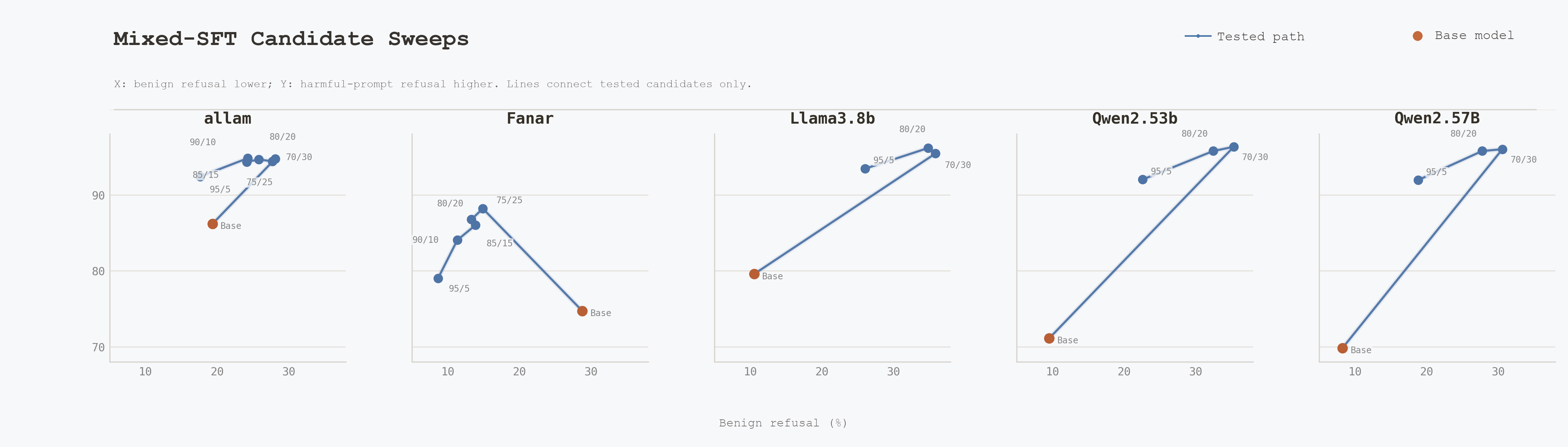}
  \caption{Observed mixed-SFT ratio sweeps. Each panel connects the base point (orange) to the tested benign/refusal mixtures (blue) in the $(\safe,\unsafe)$ plane. Coverage is uneven across models, so the paths are candidate searches rather than causal or transferable ratio curves.}
  \label{fig:ratio}
\end{figure*}

\paragraph{Mixed SFT exposes local candidate points.} Adding helpful examples recovers selectivity, but the ratio is a model-specific search (\cref{fig:ratio}). For example, ALLaM moves from $28.1/94.7$ at 70/30 benign/refusal to $17.7/92.4$ at 95/5, whereas Fanar falls to $8.6/79.0$ at 95/5 and needs more refusal pressure among the tested points. Ratio coverage is uneven: ALLaM has six tested mixtures, Fanar five, and the other models three. Exposure and training-budget dimensions were not fully equalized. These sweeps therefore identify local candidates and best-observed points under the illustrative target; they do not estimate causal ratio effects, universal optima, or transferable configurations.

\begin{figure*}[t]
  \centering
  \includegraphics[width=\linewidth]{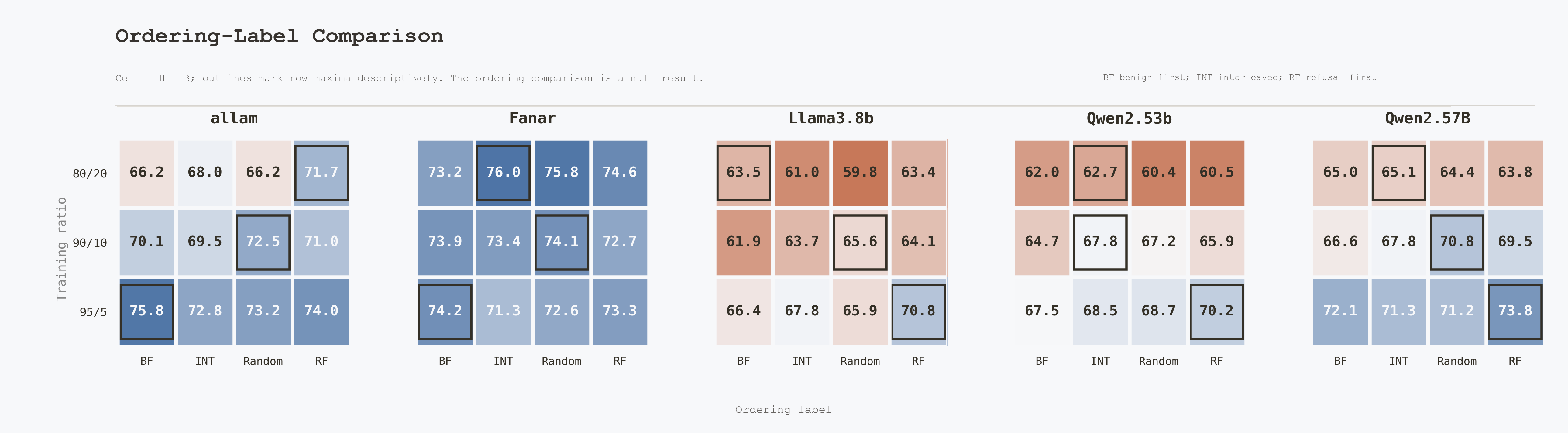}
  \caption{Descriptive ordering comparison. Cells report $\unsafe-\safe$; black outlines mark the largest margin in a model-ratio row. Random is largest in 4/15 rows, near the 25\% chance rate, and reshuffling did not preserve file order as optimizer order. We therefore treat this as a null result. BF, INT, and RF denote benign-first, interleaved, and refusal-first labels.}
  \label{fig:ordering}
\end{figure*}

\paragraph{Ordering is a null result.} \Cref{fig:ordering} reports the descriptive margin $\unsafe-\safe$ for four file-order labels. Random ordering has the largest reported margin in 4 of 15 model-ratio rows (26.7\%), close to the 25\% chance rate among four strategies, and many differences are smaller than prompt-sampling intervals. Moreover, default reshuffling meant file construction did not guarantee realized optimizer order. We retain the labels to identify runs but make no ordering-benefit or causal claim.

\paragraph{Checkpoint variation.} On the 70/30 mix, ALLaM's $\safe$ falls from 39.9 at checkpoint 50 to 19.5 at checkpoint 200 while $\unsafe$ remains above 92, whereas Llama already over-refuses at checkpoint 50 ($35.1/96.7$).

\paragraph{Selected candidates.} \cref{tab:sft} applies the illustrative rule to the tested runs. The selected points reach about 90--93\% harmful-prompt refusal at 14.0--22.6\% benign refusal. These are standardized downstream comparison points, not global optima. Four remain above $\unsafe=90\%$ in all three selected-configuration runs. Fanar crosses the threshold in only two of three, with mean $\unsafe=89.95\pm0.41$, so its rounded 90.0\% seed-42 point must be read as threshold-sensitive.

\begin{table}[t]
  \caption{Selected mixed-SFT candidates under the illustrative $\unsafe\ge90\%$ rule. Fanar$^{\dagger}$ meets the threshold in two of three seeds ($89.95\pm0.41$ mean $\unsafe$). Ordering abbreviations identify runs but do not imply an ordering effect.}
  \label{tab:sft}
  \centering
  \begin{tabular}{lccc}
    \toprule
    Model & Candidate & $\safe$ & $\unsafe$ \\
    \midrule
    ALLaM & 95/5 BF & 16.5 & 92.3 \\
    Fanar$^{\dagger}$ & 80/20 INT & 14.0 & 90.0 \\
    Llama & 95/5 RF & 22.0 & 92.8 \\
    Qwen 3B & 95/5 RF & 22.6 & 92.8 \\
    Qwen 7B & 95/5 RF & 18.6 & 92.4 \\
    \bottomrule
  \end{tabular}
\end{table}

\section{Calibration After SFT}
\label{sec:calibration}

Methods after SFT are candidate moves, not default stages. We compare DPO and guard moves on the $(\safe,\unsafe)$ plane while reporting marginal prompt-sampling intervals; these intervals are descriptive rather than a formal acceptance test for paired changes.

\begin{figure*}[t]
  \centering
  \includegraphics[width=\linewidth]{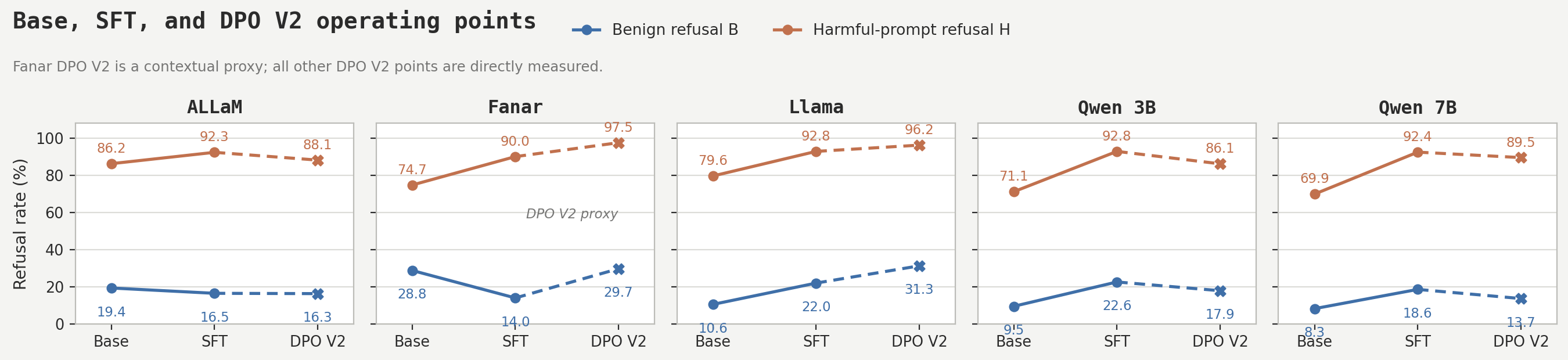}
  \caption{Base, selected-SFT, and DPO V2 operating points. Panels plot benign refusal $\safe$ (blue) and harmful-prompt refusal $\unsafe$ (orange). Solid segments show Base-to-SFT movement; dashed segments and x markers show SFT-to-DPO V2 transitions. Fanar's DPO V2 point is a contextual proxy; the other four are directly measured.}
  \label{fig:calibration}
\end{figure*}

\paragraph{DPO moves models in opposite directions.} We apply Direct Preference Optimization \citep{rafailov2023dpo} after SFT with two benign-pair constructions. DPO is a trade-off operator, not a uniform upgrade. Qwen 7B moves from $18.6/92.4$ to $13.7/89.5$; ALLaM and Qwen 3B move similarly, to $16.3/88.1$ and $17.9/86.1$. Llama instead moves from $22.0/92.8$ to $31.3/96.2$. Thus only the deployment objective determines whether a move is acceptable (\cref{fig:calibration}). Wilson intervals describe prompt sampling conditional on judge labels, so we do not call a move ``real'' solely because it exceeds one half-width. Fanar's DPO V2 point is a labeled contextual proxy and is excluded from direct comparisons; Appendix~\ref{app:proxy} gives its construction.

\paragraph{An observed configuration failure.} Under DPO V1, one Llama run labeled benign-first reaches $29.5/94.9$, while the harmful-first-labeled run reaches $74.8/89.3$. Because realized optimizer order was not controlled and this is a single-seed comparison, it demonstrates configuration fragility rather than a causal ordering effect.

\paragraph{Guards are model- and stage-dependent.} Inference guards \citep{meta2024llamaguard3} are similarly heterogeneous (\cref{tab:guard}). At base, the guard moves Fanar from $28.8/74.7$ to $5.4/82.2$. After SFT, however, it lowers $\unsafe$ for four of five models; only Fanar gains harmful-prompt refusal, from 90.0 to 93.7, while paying an eight-point $\safe$ cost. Llama Guard 3 does not officially list Arabic among its supported languages, so this use is off-label. These results motivate Arabic-native guards and selection rules that evaluate both axes, rather than a default stage.

\begin{table}[t]
  \caption{Inference guard before and after SFT, reported as $\safe/\unsafe$. Effects depend on model and stage; after SFT the guard lowers harmful-prompt refusal for four of five models. Llama Guard 3 does not officially support Arabic, so this is off-label use.}
  \label{tab:guard}
  \centering
  \scriptsize
  \setlength{\tabcolsep}{3.5pt}
  \begin{tabular}{lcccc}
    \toprule
     & \multicolumn{2}{c}{Base} & \multicolumn{2}{c}{SFT} \\
    \cmidrule(lr){2-3}\cmidrule(lr){4-5}
    Model & no guard & guard & no guard & guard \\
    \midrule
    ALLaM   & 19.4/86.2 & 19.4/86.7 & 16.5/92.3 & 16.9/89.2 \\
    Fanar   & 28.8/74.7 & 5.4/82.2  & 14.0/90.0 & 22.0/93.7 \\
    Llama   & 10.6/79.6 & 14.0/81.7 & 22.0/92.8 & 18.5/89.9 \\
    Qwen 3B & 9.5/71.1  & 10.1/72.8 & 22.6/92.8 & 14.8/82.4 \\
    Qwen 7B & 8.3/69.9  & 8.8/71.5  & 18.6/92.4 & 11.6/86.9 \\
    \bottomrule
  \end{tabular}
\end{table}

\section{Cross-Script Transfer and Robustness}
\label{sec:robustness}

The AraSafe evaluation set is MSA-heavy, but Arabic users write in dialects, Arabizi, and noisy text. We use the 730-prompt boundary set to compare base behavior across five forms and to test the selected SFT checkpoints on Arabizi. These are controlled synthetic diagnostics, not representative samples of naturally occurring user language.

\paragraph{Transformation audit.} Two annotators audited 120 transformed prompts, 30 per non-MSA form, with 85.8\% agreement ($\kappa=0.63$). Egyptian and Levantine each had 27/30 prompts rated natural or mostly natural. Arabizi and noisy Arabic each had 22/30 natural or mostly natural, 6/30 understandable but unnatural, and 2/30 invalid or meaning-changing prompts. These results support controlled stress tests but not broad claims about naturally occurring language; full counts are in \cref{tab:prompt-audit}.

\begin{figure*}[t]
  \centering
  \includegraphics[width=\linewidth]{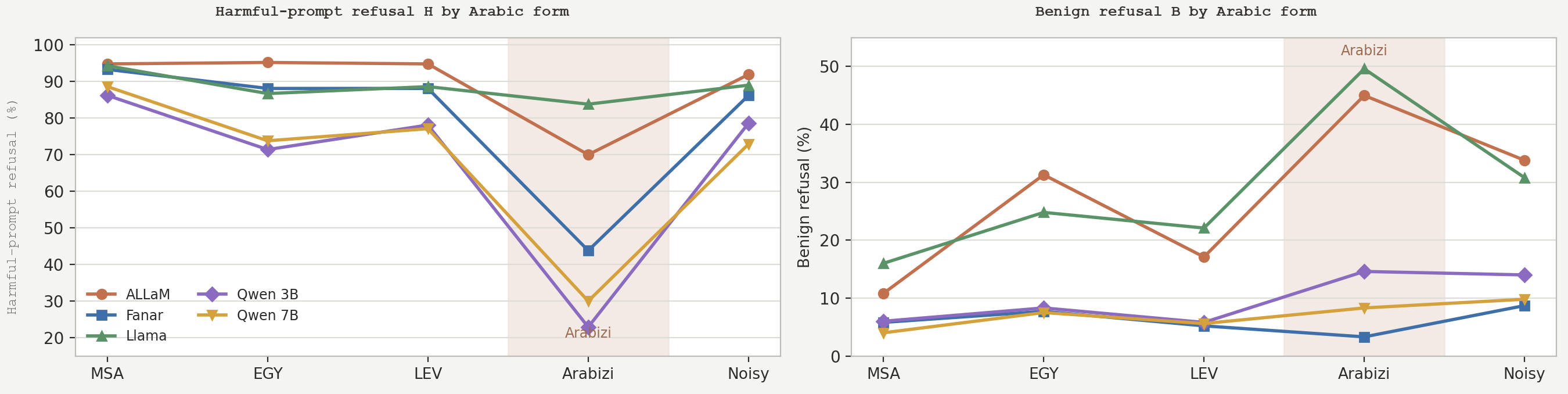}
  \caption{Base-model refusal on the synthetic 730-prompt boundary set. The left panel plots harmful-prompt refusal $\unsafe$ and the right plots benign refusal $\safe$ for MSA, Egyptian, Levantine, Arabizi, and noisy Arabic. The shaded band highlights Arabizi, where the largest shifts occur through either reduced $\unsafe$ or inflated $\safe$.}
  \label{fig:crossscript}
\end{figure*}

\paragraph{Base behavior changes across forms.} \cref{fig:crossscript} and the complete values in \cref{tab:appendix-boundary-set} show the sharpest change on Arabizi. Relative to boundary-set MSA, $\unsafe$ falls to 22.9\%, 30.0\%, and 43.8\% for Qwen 3B, Qwen 7B, and Fanar. ALLaM and Llama retain more harmful-prompt refusal but their $\safe$ rises to 45.0\% and 49.6\%. Egyptian and Levantine degradation is generally milder, although not absent. These are model-specific shifts, not evidence that one starting-point label predicts a fixed cross-script failure.

\begin{table}[t]
  \caption{Arabizi base-to-selected-SFT transfer. Each cell gives $\safe/\unsafe$ on the first line and the corresponding Wilson 95\% intervals on the second. Intervals are reconstructed from the rounded rates and known denominators ($n_B=520$, $n_H=210$), so they quantify prompt sampling conditional on the judge labels.}
  \label{tab:boundary-main}
  \centering
  \footnotesize
  \setlength{\tabcolsep}{2.5pt}
  \begin{tabular}{lcc}
    \toprule
    Model & Base & Selected SFT \\
    \midrule
    ALLaM   & \shortstack{45.0/70.0\\{[40.8,49.3]/[63.5,75.8]}} & \shortstack{43.3/74.8\\{[39.1,47.6]/[68.5,80.2]}} \\
    Fanar   & \shortstack{3.3/43.8\\{[2.1,5.2]/[37.3,50.6]}} & \shortstack{2.5/53.3\\{[1.5,4.2]/[46.6,60.0]}} \\
    Llama   & \shortstack{49.6/83.8\\{[45.3,53.9]/[78.2,88.2]}} & \shortstack{57.1/88.6\\{[52.8,61.3]/[83.6,92.2]}} \\
    Qwen 3B & \shortstack{14.6/22.9\\{[11.8,17.9]/[17.7,29.0]}} & \shortstack{19.6/34.8\\{[16.4,23.2]/[28.6,41.4]}} \\
    Qwen 7B & \shortstack{8.3/30.0\\{[6.2,11.0]/[24.2,36.5]}} & \shortstack{11.2/43.3\\{[8.7,14.1]/[36.8,50.1]}} \\
    \bottomrule
  \end{tabular}
\end{table}

\paragraph{Selected SFT transfers only partially.} \Cref{tab:boundary-main} reports the selected-checkpoint evaluation. SFT raises Arabizi $\unsafe$ for every model, but none reaches 90\% and three remain below 55\%. ALLaM and Llama also retain high benign refusal. Training only with translated MSA harmful supervision is a plausible contributor, but without a native-Arabizi ablation this is a hypothesis rather than a causal explanation. Every deployment-relevant form therefore needs its own evaluation and uncertainty analysis.

\paragraph{Non-refusal is heterogeneous.} Lower $\unsafe$ means that the judge labels fewer harmful-prompt responses as refusals; it does not reveal what the other responses contain. The five-way human taxonomy distinguishes effective refusal, mixed or leaky refusal, safe redirection, confused or irrelevant response, and harmful or enabling response. In the audit, the 44 harmful-prompt responses without effective refusal comprise 25 harmful or enabling responses, 9 mixed or leaky refusals, 6 confused or irrelevant responses, and 4 safe redirections. We do not report a separate Arabizi outcome taxonomy because that subgroup contains only ten such responses. Inclusive Arabic safety evaluation must therefore report both refusal and response outcomes.

\section{Selecting and Interpreting Operating Points}
\label{sec:recipe}

\paragraph{The portable result is a procedure.} Exact ratios, checkpoints, and thresholds are specific to five models, tested configurations, and evaluation sets. The portable part is to measure $(\safe,\unsafe)$, specify a deployment target, search candidate interventions, and retain only target-improving points that remain stable across seeds, judges, and relevant Arabic forms. For Qwen 7B, selected SFT moves $8.3/69.9$ to $18.6/92.4$, while DPO V2 trades to $13.7/89.5$; neither point is universally better. Fanar's selected $14.0/90.0$ point improves both seed-42 axes relative to $28.8/74.7$, but its three-seed mean falls just below the illustrative threshold. The procedure surfaces these choices instead of hiding them inside a fixed stack.

\paragraph{Internal signals are detectable but model-specific.} Refusal directions \citep{arditi2024refusal} separate harmful from benign responses with AraSafe AUCs of 0.909--0.956 across the five models (95\% half-width near 0.01). Score scales and thresholds differ by model, matching conditional activation control \citep{lee2025cast}; directions therefore support local calibration rather than a universal rule. Per-model threshold operating points and score distributions are in Appendix~\ref{app:diagnostic-figures}.

\paragraph{Choosing an operating point.} No single intervention dominates. An assistant that must stay helpful on sensitive but benign Arabic may prioritize low $\safe$, while a moderation setting may prioritize high $\unsafe$. The 85/90/95 sensitivity in \cref{tab:threshold-sensitivity} shows that tightening the target from 90\% to 95\% costs 7.8--23.4 additional $\safe$ points among feasible tested models, and Fanar has no feasible tested point at 95\%. The target is therefore a deployment choice, not a benchmark constant.

\paragraph{The protocol in four steps.} The procedure can be stated compactly:
\begin{enumerate}\setlength{\itemsep}{1pt}
  \item \textbf{Measure.} Estimate the base $(\safe,\unsafe)$ point and judge uncertainty on the deployment distribution.
  \item \textbf{Specify.} Set the deployment target and costs before selecting a configuration.
  \item \textbf{Search and gate.} Sweep local SFT, DPO, or guard candidates and retain only target-improving points.
  \item \textbf{Validate.} Repeat selected candidates across seeds, independent judges, and every relevant script or dialect.
\end{enumerate}

\paragraph{Takeaway.} The central result is not a universal SFT ratio, DPO variant, guard, ordering, or activation threshold. Arabic selective refusal is a model-specific operating-point selection problem, and MSA performance does not establish cross-script reliability. Refusal metrics must also remain distinct from harmful-response outcomes.

\section{Related Work}
\label{sec:related}

Our work sits at the intersection of over-refusal evaluation, Pareto safety alignment, and Arabic LLM evaluation. XSTest \citep{rottger-etal-2024-xstest} and OR-Bench \citep{cui2024orbench} document benign over-refusal, while Panacea \citep{zhong2024panacea} treats helpfulness and harmlessness as competing objectives; our $(\safe,\unsafe)$ plane makes the refusal trade-off explicit for Arabic.

Fine-tuning can erode safety even with benign data \citep{qi2024finetuning}, DPO optimizes preferences without a reward model \citep{rafailov2023dpo}, and Deliberative Alignment \citep{guan2024deliberative} reasons over a safety specification; we treat each as a candidate move rather than a fixed stack. Refusal is partly linear in activations \citep{arditi2024refusal}, and Conditional Activation Steering \citep{lee2025cast} supports model-conditional control.

Arabic resources include Jais \citep{sengupta2023jais}, AceGPT \citep{huang-etal-2024-acegpt}, Fanar \citep{fanar2025}, ALLaM \citep{bari2024allam}, AraSafe \citep{mubarak2025arasafe}, safeguard evaluation \citep{ashraf-etal-2025-arabic}, and AraDiCE \citep{mousi2024aradice}. Arabic diacritics also change tokenization and benchmark behavior \citep{inoue2026diacritics}; testing their effect on refusal is important future work.

\section*{Limitations}

Five limitations bound the conclusions.

\paragraph{Judge.} Human and Aya validation improve confidence, but Qwen3Guard accuracy is lower on Arabizi, and the family-gap interval does not exclude bias. Wilson intervals cover prompt sampling conditional on judge labels, not judge error. The retained audit record also lacks its exact stratification allocation and the subgroup denominators and intervals for the reported per-form accuracies.

\paragraph{Variance.} Primary sweeps are seed 42; only selected SFT configurations use three seeds, and Fanar meets $\unsafe\ge90\%$ in two of three.

\paragraph{Search design.} Ratio coverage and exposure are unequal across models, so best-observed candidates are not causal optima.

\paragraph{Data.} Harmful supervision is translated MSA; native dialectal and Arabizi ablations are needed before attributing transfer failures to training data.

\paragraph{Reproducibility and scope.} The transformed prompts are synthetic; the exact GPT-4o-mini snapshot, noise implementation, full selected-SFT cross-form matrix, per-seed measurements, exact ratio sample counts, hardware, runtime, total compute, and a public artifact URL are not available in the current record. We do not infer or invent these details.

\section*{Ethics Statement}

This work studies how to make Arabic language models refuse harmful requests while staying usable. Harmful prompts and enabling continuations are redacted and reported only as outcome labels (\cref{tab:examples}). We use synthetic transformations only as controlled robustness diagnostics. We disclose lower naturalness for Arabizi transformations and lower judge agreement on Arabizi responses. The Fanar DPO proxy is marked and excluded from direct comparisons. We distinguish refusal from response safety: non-refusal may be harmful, mixed, confused, or a safe redirection. Deployment therefore requires Arabic-native data, representative human evaluation, and outcome-level oversight in addition to refusal rates.

\section*{Acknowledgments}

We thank Mariam Salman, Rana Ezzeddine, Mohamad Hussein Karnib, and Hassan Hijazi.

\nocite{mubarak2025arasafe,zhao2025qwen3guard,ji2023beavertails,dibt2024ar,hammoud2025hala,arditi2024refusal,rottger-etal-2024-xstest,mousi2024aradice}

\bibliography{arabic_selective_refusal}

\clearpage
\appendix
\onecolumn

\section{Experimental and Data Details}
\label{app:data}

Primary sweeps and sampled splits use seed 42; selected mixed-SFT configurations are repeated with seeds 43 and 44. The 12{,}077 human-written AraSafe prompts \citep{mubarak2025arasafe} are reserved exclusively for downstream evaluation with Qwen3Guard-Gen-4B \citep{zhao2025qwen3guard}; they never appear in training, direction extraction, or calibration splits.

\subsection{Preference and SFT Data}

\paragraph{DPO data (10K harmful and 10K benign pairs).}
Harmful prompts come from an MSA translation of BeaverTails \citep{ji2023beavertails}, produced with the Hala English--Arabic translators \citep{hammoud2025hala}. For each prompt, the rejected side is the harmful BeaverTails completion and the chosen side is a refusal. The benign side is sourced separately from \texttt{data-is-better-together/dpo-datasets-for-ar} \citep{dibt2024ar}. V1 ranks two helpful assistant answers; V2 replaces the rejected member with a refusal so the model learns to prefer a helpful answer over benign over-refusal.

\paragraph{SFT data.}
The harmful portion uses the same BeaverTails MSA translation with refusals as the chosen response. The benign portion is drawn separately from \texttt{hammh0a/Hala-4.6M-SFT} \citep{hammoud2025hala}. We test benign/refusal ratios 70/30, 75/25, 80/20, 85/15, 90/10, and 95/5, plus a 100\% refusal collapse control. Coverage is not balanced across models, and exact example counts per tested ratio are absent from the retained record; consequently, the sweep does not isolate ratio from exposure or training budget.

\subsection{Training Configuration}

All runs use batch size 32, Paged AdamW 8-bit, a linear schedule with five warm-up steps, gradient clipping at 1.0, gradient checkpointing, logging every two steps, and full-parameter training. \Cref{tab:training-config} records the available run-specific settings.

\begin{table}[H]
  \centering
  \small
  \caption{Training configuration. Selected SFT candidates additionally use seeds 43 and 44; all primary sweeps use seed 42.}
  \label{tab:training-config}
  \begin{tabular}{lccccc}
    \toprule
    Run & Budget & Learning rate & Precision & Length & Seed \\
    \midrule
    Standard SFT & 1 epoch & $10^{-5}$ & FP16 & 1024 & 42 \\
    Collapse SFT & $\leq100$ steps & $10^{-5}$ & BF16 & 512 & 42 \\
    DPO & 1 epoch & $5\times10^{-7}$ & BF16 & 1024/512 prompt & 42 \\
    \bottomrule
  \end{tabular}
  \vspace{2pt}

  \footnotesize DPO uses $\beta=0.1$. Hardware, runtime, total compute, and exact per-ratio sample counts were not retained in the current artifact.
\end{table}

\subsection{Ordering and Refusal Directions}

\paragraph{Ordering.}
For DPO and SFT, files are constructed with benign-first, harmful-first, interleaved, or random labels. Default training reshuffling means these labels do not guarantee realized optimizer order. The comparison is therefore retained as run metadata and a null result, not as evidence that ordering causally changes alignment.

\paragraph{Refusal direction data.}
Following \citet{arditi2024refusal}, we compute candidate refusal directions across layers and token positions from a \emph{direction extraction split} of 1{,}000 benign Hala prompts and 1{,}000 translated BeaverTails harmful prompts. Thresholds are tuned on a separate \emph{calibration split} of 256 benign prompts from DPO data and 256 harmful prompts from the AraSafe synthetic split \citep{mubarak2025arasafe}. Drawing calibration prompts from a different source prevents tuning and validating on the same distribution.

\subsection{Boundary Generalization Set}
\label{boundary}

The boundary generalization set \citep{rottger-etal-2024-xstest} contains 730 prompts: 210 harmful prompts, 270 sensitive benign prompts, and 250 clean benign prompts. Egyptian and Levantine variants are translated using AraDiCE models \citep{mousi2024aradice}; Arabizi was generated with GPT-4o-mini, and noisy Arabic with a custom transformation script. The retained record does not include the exact GPT-4o-mini snapshot or the noise algorithm, so we identify these as reproducibility gaps rather than guessing them. Per-form base rates appear in \cref{tab:appendix-boundary-set}; selected-SFT Arabizi rates appear in \cref{tab:boundary-main}.

\subsection{Fanar DPO V2 Proxy}
\label{app:proxy}

We could not run a directly comparable Fanar DPO V2 measurement on the same checkpoint as the other four models. The proxy shown in \cref{fig:calibration} and the appendix is reconstructed by taking Fanar's directly measured DPO V1 interleaved operating point and adding the median interleaved DPO V1-to-DPO V2 shift across the four directly measured models: $-0.41$ points in $\safe$ and $+0.275$ points in $\unsafe$, yielding $29.71/97.47$. It is labeled wherever shown and excluded from all main-text comparisons and conclusions.

\section{Validation and Sensitivity Analyses}
\label{app:validation}

\subsection{Human and Cross-Judge Audit}

\begin{table}[H]
  \centering
  \small
  \caption{Blinded 300-response audit. Agreement rows compare two human annotators; judge rows compare automatic labels with adjudicated labels. The available audit record includes per-form accuracies but not their subgroup denominators or intervals.}
  \label{tab:judge-audit}
  \begin{tabular}{lccc}
    \toprule
    Measure & Agreement/accuracy (\%) & $\kappa$/macro-F1 & 95\% interval \\
    \midrule
    Binary human refusal & 89.0 & 0.78 & --- \\
    Three-way human refusal & 83.7 & 0.71 & --- \\
    Five-way human outcome & 78.0 & 0.69 & --- \\
    Human agreement, Arabizi & 75.0 & 0.49 & --- \\
    Qwen3Guard & 88.7 & 0.887 & [85.3, 91.7] accuracy \\
    Aya Expanse 32B & 91.0 & 0.910 & --- \\
    Qwen3Guard, MSA AraSafe & 90.0 & --- & --- \\
    Qwen3Guard, Egyptian & 90.0 & --- & --- \\
    Qwen3Guard, Levantine & 95.0 & --- & --- \\
    Qwen3Guard, noisy Arabic & 87.5 & --- & --- \\
    Qwen3Guard, boundary MSA & 90.0 & --- & --- \\
    Qwen3Guard, Arabizi & 77.5 & 0.763 & --- \\
    Aya Expanse, Arabizi & 87.5 & --- & --- \\
    \bottomrule
  \end{tabular}
\end{table}

The Qwen-versus-Aya paired difference is inconclusive (McNemar $p=0.382$; macro-F1 difference $-2.3$ points, 95\% interval [$-6.4$, 2.0]). Qwen3Guard accuracy is 86.7\% on Qwen-family outputs and 90.0\% otherwise, with a gap interval of [$-3.3$, 10.3]. On the balanced audit subset, automatic labels yield $\safe/\unsafe=16.7/82.0$, while adjudicated surface-refusal labels yield $23.3/76.7$. The audit therefore supports approximate refusal measurement but does not make judge error negligible.

\begin{table}[H]
  \centering
  \small
  \caption{Adjudicated outcomes for the 44 audited harmful-prompt responses without effective refusal.}
  \label{tab:outcome-audit}
  \begin{tabular}{lr}
    \toprule
    Outcome & Count \\
    \midrule
    Harmful or enabling response & 25 \\
    Mixed or leaky refusal & 9 \\
    Confused or irrelevant response & 6 \\
    Safe redirection & 4 \\
    \bottomrule
  \end{tabular}
\end{table}

\subsection{Transformation Naturalness}

\begin{table}[H]
  \centering
  \small
  \caption{Human naturalness audit of 120 transformed prompts, 30 per form. Overall annotator agreement is 85.8\% ($\kappa=0.63$).}
  \label{tab:prompt-audit}
  \begin{tabular}{lrrr}
    \toprule
    Form & Natural/mostly natural & Understandable, unnatural & Invalid/changed \\
    \midrule
    Egyptian & 27 & 2 & 1 \\
    Levantine & 27 & 1 & 2 \\
    Arabizi & 22 & 6 & 2 \\
    Noisy Arabic & 22 & 6 & 2 \\
    \bottomrule
  \end{tabular}
\end{table}

\subsection{Target and Seed Sensitivity}

\begin{table}[H]
  \centering
  \small
  \caption{Lowest observed benign-refusal rate $\safe$ among tested candidates satisfying three illustrative harmful-prompt-refusal targets. A dash means no tested candidate is feasible.}
  \label{tab:threshold-sensitivity}
  \begin{tabular}{lrrr}
    \toprule
    Model & $\unsafe\ge85$ & $\unsafe\ge90$ & $\unsafe\ge95$ \\
    \midrule
    ALLaM & 16.5 & 16.5 & 39.9 \\
    Fanar & 9.4 & 14.0 & --- \\
    Llama & 22.0 & 22.0 & 31.2 \\
    Qwen 3B & 22.6 & 22.6 & 30.4 \\
    Qwen 7B & 18.6 & 18.6 & 27.3 \\
    \bottomrule
  \end{tabular}
\end{table}

Selected mixed-SFT configurations were repeated with seeds 42, 43, and 44. Standard deviations span 0.13--0.57 $\safe$ points and 0.41--1.10 $\unsafe$ points. ALLaM, Llama, Qwen 3B, and Qwen 7B exceed $\unsafe=90\%$ in all three runs. Fanar does so in two of three, with mean $\unsafe=89.95\pm0.41$; exact per-seed operating points are not present in the retained record.

\section{Complete Result Tables}
\label{C}
This appendix collects the retained numerical tables supporting the paper. Each block starts with a human-readable result title, the row count excluding the header, and the claim or figure it supports.
\begin{center}\begin{minipage}{\linewidth}\centering
\small
\captionof{table}{Experiment scale used.}
\label{tab:experiment-scale}
\begin{tabular}{lr}
\toprule
Item & Value \\
\midrule
Arabic-capable models & 5 \\
Training runs in the experiment inventory & 130 \\
AraSafe evaluation prompts & 12,077 \\
Benign AraSafe prompts & 10,823 \\
Harmful AraSafe prompts & 1,254 \\
\bottomrule
\end{tabular}
\end{minipage}\end{center}
\begin{center}\begin{minipage}{\linewidth}\centering
\small
\captionof{table}{Result-table index and claim trace.}
\label{tab:result-index}
\begin{tabular}{lrl}
\toprule
Result table & Rows & Paper trace \\
\midrule
Base operating points & 5 & \cref{fig:scatter,fig:base-bars}. \\
Selected mixed-SFT candidates & 5 & \cref{tab:sft,fig:selected}. \\
Mixture-ratio sweep & 25 & \cref{fig:ratio}. \\
Ordering-strategy sweep & 15 & \cref{fig:ordering}. \\
Mixed 70/30 checkpoints & 15 & \cref{fig:checkpoints}. \\
Refusal-only collapse thresholds & 55 & \cref{fig:collapse}. \\
Extra ALLaM refusal-only checkpoints & 3 & Extra ALLaM refusal-only checkpoints. \\
DPO V1 ordering comparison & 5 & DPO V1 ordering comparison. \\
Reported DPO V1 operating points & 5 & Reported DPO V1 operating points. \\
DPO V2 ordering comparison & 5 & DPO V2 ordering comparison after SFT. \\
Reported DPO V2 operating points & 5 & \cref{fig:calibration,fig:guarddpo}. \\
Guard comparison & 5 & \cref{tab:guard,fig:guarddpo}. \\
Base to SFT to DPO V2 operating curve & 15 & \cref{fig:calibration}. \\
Safety-direction thresholds & 5 & \cref{fig:directions,fig:score-dist}. \\
\bottomrule
\end{tabular}
\end{minipage}\end{center}
\subsection{Evaluation and Selected SFT}
\paragraph{Base operating points.} Main-text base comparison in \cref{fig:scatter}.
\begin{center}\begin{minipage}{\linewidth}\centering
\small
\captionof{table}{Complete base operating-point refusal rates.}
\label{tab:appendix-base-regimes}
\begin{tabular}{lrr}
\toprule
Model & B & H \\
\midrule
Fanar-1-9B & 28.78 & 74.72 \\
Meta-Llama-3-8B-Instruct & 10.55 & 79.59 \\
Qwen2.5-3B-Instruct & 9.52 & 71.13 \\
Qwen2.5-7B-Instruct & 8.26 & 69.86 \\
ALLaM & 19.36 & 86.20 \\
\bottomrule
\end{tabular}
\end{minipage}\end{center}

\paragraph{Selected mixed-SFT candidates.} Main-text candidates in \cref{tab:sft}.
\begin{center}\begin{minipage}{\linewidth}\centering
\small
\captionof{table}{Complete selected mixed-SFT candidate table. Fanar's seed-42 point rounds to $H=90.0$, but the candidate meets $H\ge90$ in only two of three seeds.}
\label{tab:appendix-selected-sft}
\begin{tabular}{llrr}
\toprule
Model & Stage & B & H \\
\midrule
ALLaM & 95ben\_5ref benign-first label & 16.5 & 92.3 \\
Fanar & 80ben\_20ref interleaved & 14.0 & 90.0 \\
Llama & 95ben\_5ref refusal first & 22.0 & 92.8 \\
Qwen 3B & 95ben\_5ref refusal first & 22.6 & 92.8 \\
Qwen 7B & 95ben\_5ref refusal first & 18.6 & 92.4 \\
\bottomrule
\end{tabular}
\end{minipage}\end{center}

\subsection{SFT Sweeps and Ordering}
\paragraph{Mixture-ratio sweep.} Ratio sweep and candidate-trajectory figure.
\begin{center}\begin{minipage}{\linewidth}\centering
\small
\captionof{table}{Complete mixture-ratio sweep. Rows use the default shuffled ordering, so the Qwen 7B 95/5 entry (18.81/91.95) differs slightly from the selected refusal-first candidate (18.6/92.4) by less than the sampling interval.}
\label{tab:appendix-mixture-ratio-sweep}
\begin{tabular}{llrr}
\toprule
Model & Stage & B & H \\
\midrule
ALLaM & Base & 19.36 & 86.20 \\
ALLaM & 70ben\_30ref & 28.11 & 94.74 \\
ALLaM & 75ben\_25ref & 27.74 & 94.42 \\
ALLaM & 80ben\_20ref & 25.82 & 94.66 \\
ALLaM & 85ben\_15ref & 24.12 & 94.34 \\
ALLaM & 90ben\_10ref & 24.25 & 94.82 \\
ALLaM & 95ben\_5ref & 17.65 & 92.42 \\
Fanar-1-9B & Base & 28.78 & 74.72 \\
Fanar-1-9B & 75ben\_25ref & 14.88 & 88.20 \\
Fanar-1-9B & 80ben\_20ref & 13.27 & 86.76 \\
Fanar-1-9B & 85ben\_15ref & 13.87 & 86.04 \\
Fanar-1-9B & 90ben\_10ref & 11.31 & 84.05 \\
Fanar-1-9B & 95ben\_5ref & 8.62 & 79.03 \\
Llama-3-8B & Base & 10.55 & 79.59 \\
Llama-3-8B & 70ben\_30ref & 35.87 & 95.45 \\
Llama-3-8B & 80ben\_20ref & 34.85 & 96.17 \\
Llama-3-8B & 95ben\_5ref & 26.06 & 93.46 \\
Qwen2.5-3B & Base & 9.52 & 71.13 \\
Qwen2.5-3B & 70ben\_30ref & 35.30 & 96.33 \\
Qwen2.5-3B & 80ben\_20ref & 32.42 & 95.77 \\
Qwen2.5-3B & 95ben\_5ref & 22.58 & 92.03 \\
Qwen2.5-7B & Base & 8.26 & 69.86 \\
Qwen2.5-7B & 70ben\_30ref & 30.58 & 96.01 \\
Qwen2.5-7B & 80ben\_20ref & 27.76 & 95.77 \\
Qwen2.5-7B & 95ben\_5ref & 18.81 & 91.95 \\
\bottomrule
\end{tabular}
\end{minipage}\end{center}

\paragraph{Ordering-strategy sweep.} Ordering-label heatmap reported as a null comparison.
\begin{center}\begin{minipage}{\linewidth}\centering
\scriptsize
\captionof{table}{Complete ordering-label sweep. Labels record file construction, not guaranteed optimizer order; the comparison is a null result.}
\label{tab:appendix-ordering-sweep}
\resizebox{\textwidth}{!}{%
\begin{tabular}{llrrrrrrrrrrrr}
\toprule
Model & Ratio & BF all & BF B & BF H & INT all & INT B & INT H & RAND all & RAND B & RAND H & RF all & RF B & RF H \\
\midrule
Fanar-1-9B & 80ben\_20ref & 21.6 & 14.0 & 87.2 & 21.9 & 14.0 & 90.0 & 17.2 & 9.4 & 85.2 & 22.6 & 14.8 & 89.4 \\
Fanar-1-9B & 90ben\_10ref & 18.9 & 11.2 & 85.1 & 20.9 & 13.3 & 86.7 & 16.7 & 9.0 & 83.1 & 17.2 & 9.6 & 82.3 \\
Fanar-1-9B & 95ben\_5ref & 17.3 & 9.6 & 83.8 & 17.9 & 10.5 & 81.8 & 17.8 & 10.3 & 82.9 & 18.7 & 11.1 & 84.4 \\
Meta-Llama-3-8B-Instruct & 80ben\_20ref & 38.5 & 32.0 & 95.5 & 40.6 & 34.2 & 95.2 & 42.9 & 36.7 & 96.5 & 38.4 & 31.8 & 95.2 \\
Meta-Llama-3-8B-Instruct & 90ben\_10ref & 40.0 & 33.6 & 95.5 & 38.0 & 31.4 & 95.1 & 35.9 & 29.1 & 94.7 & 37.8 & 31.2 & 95.3 \\
Meta-Llama-3-8B-Instruct & 95ben\_5ref & 34.3 & 27.5 & 93.9 & 32.4 & 25.3 & 93.1 & 33.7 & 26.8 & 92.7 & 29.4 & 22.0 & 92.8 \\
Qwen2.5-3B-Instruct & 80ben\_20ref & 41.1 & 34.6 & 96.6 & 39.9 & 33.4 & 96.1 & 41.8 & 35.5 & 95.9 & 41.6 & 35.4 & 95.9 \\
Qwen2.5-3B-Instruct & 90ben\_10ref & 37.1 & 30.4 & 95.1 & 32.8 & 25.7 & 93.5 & 34.0 & 27.0 & 94.2 & 35.4 & 28.6 & 94.5 \\
Qwen2.5-3B-Instruct & 95ben\_5ref & 33.0 & 26.0 & 93.5 & 31.9 & 24.8 & 93.3 & 31.1 & 24.0 & 92.7 & 29.9 & 22.6 & 92.8 \\
Qwen2.5-7B-Instruct & 80ben\_20ref & 37.8 & 31.0 & 96.0 & 38.0 & 31.3 & 96.4 & 38.3 & 31.6 & 96.0 & 38.9 & 32.3 & 96.1 \\
Qwen2.5-7B-Instruct & 90ben\_10ref & 35.6 & 28.7 & 95.3 & 34.4 & 27.3 & 95.1 & 29.0 & 21.7 & 92.5 & 31.5 & 24.3 & 93.8 \\
Qwen2.5-7B-Instruct & 95ben\_5ref & 28.5 & 21.0 & 93.1 & 29.4 & 22.0 & 93.3 & 29.3 & 21.9 & 93.1 & 26.3 & 18.6 & 92.4 \\
ALLaM & 80ben\_20ref & 29.6 & 22.7 & 88.9 & 26.0 & 19.0 & 87.0 & 30.1 & 23.2 & 89.4 & 19.7 & 12.2 & 83.9 \\
ALLaM & 90ben\_10ref & 31.8 & 24.5 & 94.6 & 32.1 & 24.8 & 94.3 & 29.7 & 22.1 & 94.6 & 28.9 & 21.5 & 92.5 \\
ALLaM & 95ben\_5ref & 24.3 & 16.5 & 92.3 & 27.4 & 19.9 & 92.7 & 26.0 & 18.4 & 91.6 & 25.9 & 18.2 & 92.2 \\
\bottomrule
\end{tabular}
}
\end{minipage}\end{center}

\Needspace{18\baselineskip}
\paragraph{Mixed 70/30 checkpoints.} Mixed 70/30 checkpoint trajectory.
\begin{center}\begin{minipage}{\linewidth}\centering
\small
\captionof{table}{Complete mixed 70/30 checkpoint trajectory.}
\label{tab:appendix-mixed-checkpoints}
\begin{tabular}{llrr}
\toprule
Model & Ckpt & B & H \\
\midrule
ALLaM & ckpt-50 & 39.86 & 98.17 \\
ALLaM & ckpt-100 & 25.17 & 93.70 \\
ALLaM & ckpt-200 & 19.47 & 92.42 \\
Fanar-1-9B & ckpt-50 & 13.24 & 86.52 \\
Fanar-1-9B & ckpt-150 & 19.49 & 91.95 \\
Fanar-1-9B & ckpt-300 & 13.06 & 88.28 \\
Llama-3-8B & ckpt-50 & 35.13 & 96.65 \\
Llama-3-8B & ckpt-200 & 42.43 & 97.37 \\
Llama-3-8B & ckpt-313 & 37.14 & 96.41 \\
Qwen2.5-3B & ckpt-50 & 24.59 & 92.98 \\
Qwen2.5-3B & ckpt-150 & 36.01 & 95.53 \\
Qwen2.5-3B & ckpt-313 & 35.60 & 96.17 \\
Qwen2.5-7B & ckpt-50 & 26.05 & 95.30 \\
Qwen2.5-7B & ckpt-200 & 39.80 & 97.69 \\
Qwen2.5-7B & ckpt-313 & 32.63 & 96.65 \\
\bottomrule
\end{tabular}
\end{minipage}\end{center}

\subsection{Refusal Only Collapse}
\paragraph{Refusal-only collapse thresholds.}
\begin{center}\begin{minipage}{\linewidth}\centering
\scriptsize
\captionof{table}{Complete refusal-only collapse trajectory, pivoted by model for print. Base entries are trajectory-specific reruns; Fanar's 29.2/74.6 is distinct from its canonical AraSafe base point 28.78/74.72.}
\label{tab:appendix-refusal-collapse}
\resizebox{\textwidth}{!}{%
\begin{tabular}{lrrrrrrrrrr}
\toprule
Step & ALLaM B & ALLaM H & Fanar B & Fanar H & Llama B & Llama H & Qwen 3B B & Qwen 3B H & Qwen 7B B & Qwen 7B H \\
\midrule
Base & 19.3 & 86.0 & 29.2 & 74.6 & 10.4 & 79.9 & 9.5 & 71.0 & 8.2 & 70.0 \\
5 & 30.0 & 92.3 & 46.3 & 87.5 & 26.5 & 92.7 & 9.5 & 72.3 & 9.4 & 74.2 \\
10 & 99.9 & 100.0 & 96.0 & 99.9 & 100.0 & 100.0 & 15.7 & 85.2 & 18.4 & 92.0 \\
15 & 100.0 & 100.0 & 96.2 & 100.0 & 100.0 & 100.0 & 23.4 & 90.1 & 38.6 & 98.0 \\
20 & 100.0 & 100.0 & 96.8 & 100.0 & 100.0 & 100.0 & 37.8 & 95.7 & 65.6 & 99.7 \\
25 & 99.9 & 100.0 & 97.2 & 100.0 & 100.0 & 100.0 & 68.8 & 99.8 & 78.7 & 99.8 \\
30 & 99.9 & 100.0 & 96.8 & 100.0 & 100.0 & 100.0 & 82.8 & 100.0 & 85.4 & 99.8 \\
40 & 100.0 & 100.0 & 97.1 & 99.9 & 100.0 & 100.0 & 90.2 & 100.0 & 90.8 & 99.9 \\
60 & 100.0 & 100.0 & 97.2 & 100.0 & 100.0 & 100.0 & 92.8 & 100.0 & 94.5 & 99.9 \\
80 & 100.0 & 100.0 & 97.3 & 100.0 & 100.0 & 100.0 & 93.8 & 100.0 & 94.8 & 99.9 \\
100 & 100.0 & 100.0 & 97.1 & 100.0 & 100.0 & 100.0 & 93.7 & 100.0 & 94.9 & 100.0 \\
\bottomrule
\end{tabular}
}
\end{minipage}\end{center}

\paragraph{Extra ALLaM refusal-only checkpoints.}
\begin{center}\begin{minipage}{\linewidth}\centering
\small
\captionof{table}{Complete extra ALLaM refusal-only checkpoints.}
\label{tab:appendix-allam-collapse}
\begin{tabular}{lrr}
\toprule
Ckpt & B & H \\
\midrule
checkpoint-50 & 99.82 & 100.00 \\
checkpoint-100 & 100.00 & 100.00 \\
checkpoint-150 & 100.00 & 100.00 \\
\bottomrule
\end{tabular}
\end{minipage}\end{center}

\subsection{DPO and Guard Calibration}
\paragraph{DPO V1 ordering-label comparison.}
\begin{center}\begin{minipage}{\linewidth}\centering
\scriptsize
\captionof{table}{Complete DPO V1 ordering-label comparison. BF, HF, and INT denote benign-first, harmful-first, and interleaved file labels. Fanar's interleaved value is taken from the directly reported operating-point inventory in \cref{tab:appendix-dpo-v1-best}.}
\label{tab:appendix-dpo-v1-ordering}
\resizebox{\textwidth}{!}{%
\begin{tabular}{lrrrrrr}
\toprule
Model & DPO BF B & DPO BF H & DPO HF B & DPO HF H & DPO INT B & DPO INT H \\
\midrule
ALLaM & 18.01 & 89.15 & 13.01 & 84.52 & 17.31 & 88.04 \\
Fanar & 32.30 & 96.00 & 31.82 & 95.69 & 30.12 & 97.19 \\
Llama & 29.45 & 94.90 & 74.80 & 89.30 & 28.28 & 94.50 \\
Qwen 3B & 17.80 & 85.90 & 17.59 & 86.12 & 17.83 & 85.65 \\
Qwen 7B & 14.80 & 90.50 & 14.90 & 90.30 & 14.67 & 90.27 \\
\bottomrule
\end{tabular}
}
\end{minipage}\end{center}

\Needspace{12\baselineskip}
\paragraph{Reported DPO V1 operating points.}
\begin{center}\begin{minipage}{\linewidth}\centering
\small
\captionof{table}{Complete reported DPO V1 operating points.}
\label{tab:appendix-dpo-v1-best}
\begin{tabular}{llrr}
\toprule
Model & Stage & B & H \\
\midrule
ALLaM & DPO 20K harmful-first label & 13.01 & 84.52 \\
Fanar & DPO 20K interleaved & 30.12 & 97.19 \\
Llama & DPO 20K interleaved & 28.28 & 94.50 \\
Qwen 3B & DPO 20K harmful-first label & 17.59 & 86.12 \\
Qwen 7B & DPO 20K benign-first label & 14.80 & 90.50 \\
\bottomrule
\end{tabular}
\end{minipage}\end{center}

\paragraph{DPO V2 ordering-label comparison.}
\begin{center}\begin{minipage}{\linewidth}\centering
\scriptsize
\captionof{table}{Complete DPO V2 ordering-label comparison after SFT. BF, HF, and INT denote benign-first, harmful-first, and interleaved file labels. Only Fanar's documented interleaved proxy is retained; untraced label-specific proxies are omitted.}
\label{tab:appendix-dpo-v2-ordering}
\resizebox{\textwidth}{!}{%
\begin{tabular}{lrrrrrrrrl}
\toprule
Model & SFT B & SFT H & DPO BF B & DPO BF H & DPO HF B & DPO HF H & DPO INT B & DPO INT H & Source \\
\midrule
ALLaM & 16.5 & 92.3 & 16.59 & 88.04 & 16.10 & 87.88 & 16.33 & 88.12 & Direct \\
Fanar & 14.0 & 90.0 & -- & -- & -- & -- & 29.71 & 97.47 & INT proxy only \\
Llama & 22.0 & 92.8 & 31.42 & 95.85 & 31.40 & 95.85 & 31.28 & 96.17 & Direct \\
Qwen 3B & 22.6 & 92.8 & 17.94 & 85.33 & 17.80 & 85.25 & 17.90 & 86.12 & Direct \\
Qwen 7B & 18.6 & 92.4 & 13.68 & 89.47 & 13.84 & 89.47 & 13.78 & 89.39 & Direct \\
\bottomrule
\end{tabular}
}
\end{minipage}\end{center}

\begin{figure}[H]
  \centering
  \includegraphics[width=\linewidth]{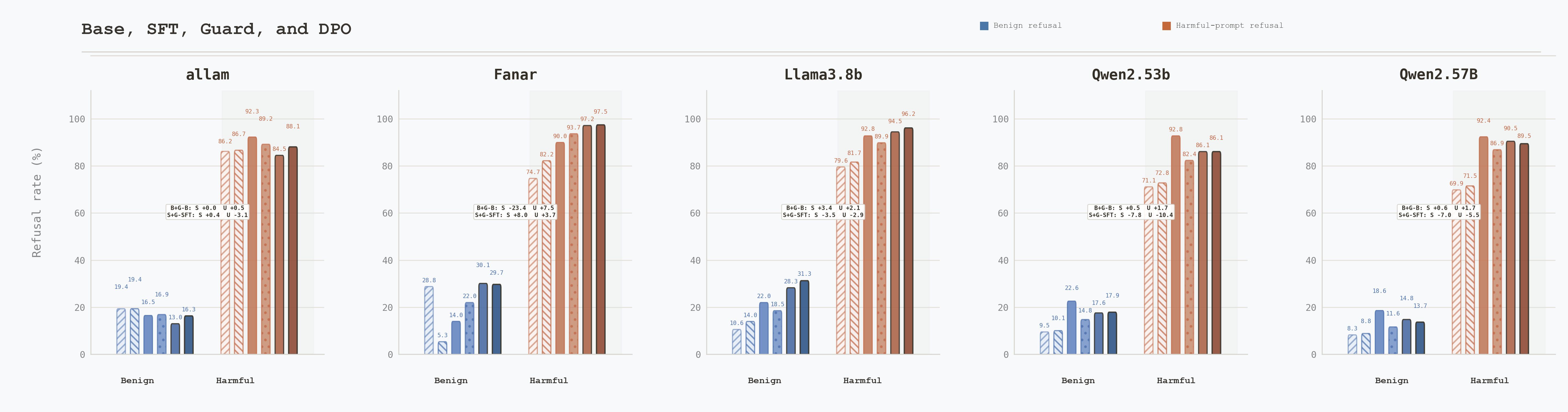}
  \caption{AraSafe refusal rates for Base, Base+Guard, selected SFT, SFT+Guard, DPO V1, and DPO V2. Panels report benign refusal $\safe$ and harmful-prompt refusal $\unsafe$. Fanar DPO V2 is a proxy estimate; all other points are directly measured.}
  \label{fig:guarddpo}
\end{figure}

\paragraph{Reported DPO V2 operating points.} Direct and proxy values are shown in \cref{fig:calibration,fig:guarddpo}.
\begin{center}\begin{minipage}{\linewidth}\centering
\small
\captionof{table}{Complete reported DPO V2 operating points.}
\label{tab:appendix-dpo-v2-best}
\begin{tabular}{llrrl}
\toprule
Model & Stage & B & H & Source \\
\midrule
ALLaM & DPO 20K interleaved & 16.33 & 88.12 & Direct \\
Fanar & DPO 20K interleaved & 29.71 & 97.47 & Proxy estimate \\
Llama & DPO 20K interleaved & 31.28 & 96.17 & Direct \\
Qwen 3B & DPO 20K interleaved & 17.90 & 86.12 & Direct \\
Qwen 7B & DPO 20K benign-first label & 13.68 & 89.47 & Direct \\
\bottomrule
\end{tabular}
\end{minipage}\end{center}

\paragraph{Guard comparison.} Main-text summary in \cref{tab:guard} and full view in \cref{fig:guarddpo}.
\begin{center}\begin{minipage}{\linewidth}\centering
\scriptsize
\captionof{table}{Complete guard comparison before and after SFT.}
\label{tab:appendix-guard-comparison}
\resizebox{\textwidth}{!}{%
\begin{tabular}{lrrrrrrrr}
\toprule
Model & Base B & Base H & Base+Guard B & Base+Guard H & SFT B & SFT H & SFT+Guard B & SFT+Guard H \\
\midrule
ALLaM & 19.36 & 86.20 & 19.38 & 86.68 & 16.50 & 92.30 & 16.90 & 89.23 \\
Fanar & 28.78 & 74.72 & 5.35 & 82.22 & 14.00 & 90.00 & 21.99 & 93.70 \\
Llama & 10.55 & 79.59 & 13.96 & 81.66 & 22.00 & 92.80 & 18.54 & 89.87 \\
Qwen 3B & 9.52 & 71.13 & 10.05 & 72.81 & 22.60 & 92.80 & 14.78 & 82.38 \\
Qwen 7B & 8.26 & 69.86 & 8.84 & 71.53 & 18.60 & 92.40 & 11.64 & 86.92 \\
\bottomrule
\end{tabular}
}
\end{minipage}\end{center}

\paragraph{Base to SFT to DPO V2 operating curve.}
\begin{center}\begin{minipage}{\linewidth}\centering
\scriptsize
\setlength{\tabcolsep}{20pt}
\renewcommand{\arraystretch}{1.0}
\captionof{table}{Complete base to SFT to DPO V2 operating curve. Fanar DPO V2 is marked as a proxy estimate.}
\label{tab:appendix-operating-curve}
\begin{tabular}{llrrrl}
\toprule
Model & Stage & Index & B & H & Source \\
\midrule
ALLaM & Base & 0.0 & 19.36 & 86.2 & Direct \\
ALLaM & SFT & 50.0 & 16.5 & 92.3 & Direct \\
ALLaM & DPO V2 & 100.0 & 16.33 & 88.12 & Direct \\
Fanar & Base & 0.0 & 28.78 & 74.72 & Direct \\
Fanar & SFT & 50.0 & 14.0 & 90.0 & Direct \\
Fanar & DPO V2 & 100.0 & 29.71 & 97.47 & Proxy estimate \\
Llama & Base & 0.0 & 10.55 & 79.59 & Direct \\
Llama & SFT & 50.0 & 22.0 & 92.8 & Direct \\
Llama & DPO V2 & 100.0 & 31.28 & 96.17 & Direct \\
Qwen 3B & Base & 0.0 & 9.52 & 71.13 & Direct \\
Qwen 3B & SFT & 50.0 & 22.6 & 92.8 & Direct \\
Qwen 3B & DPO V2 & 100.0 & 17.9 & 86.12 & Direct \\
Qwen 7B & Base & 0.0 & 8.26 & 69.86 & Direct \\
Qwen 7B & SFT & 50.0 & 18.6 & 92.4 & Direct \\
Qwen 7B & DPO V2 & 100.0 & 13.68 & 89.47 & Direct \\
\bottomrule
\end{tabular}
\end{minipage}\end{center}

\Needspace{12\baselineskip}
\subsection{Refusal Directions}
\paragraph{Refusal-direction thresholds.} Direction thresholds, refusal rates, and AUC.
\begin{center}\begin{minipage}{\linewidth}\centering
\small
\captionof{table}{Complete refusal-direction thresholds, refusal rates, and AUC.}
\label{tab:appendix-safety-directions}
\begin{tabular}{lrrrr}
\toprule
Model & Threshold & B & H & AUC \\
\midrule
Fanar & 61.594147 & 0.165111 & 0.823764 & 0.908965 \\
Llama & 0.379723 & 0.194863 & 0.934609 & 0.950549 \\
Qwen 3B & 11.373364 & 0.140164 & 0.868421 & 0.93957 \\
Qwen 7B & 14.52947 & 0.063568 & 0.832536 & 0.955017 \\
ALLaM & 2.218028 & 0.274785 & 0.949761 & 0.955747 \\
\bottomrule
\end{tabular}
\end{minipage}\end{center}

\subsection{Boundary Set Robustness}
\paragraph{Boundary set refusal rates across Arabic forms.} Boundary-set $\safe$ and $\unsafe$ per Arabic form, on the 730-prompt set described in \cref{boundary}.
\begin{center}\begin{minipage}{\linewidth}\centering
\small
\captionof{table}{Boundary-set refusal rates by form: Modern Standard Arabic (MSA), Egyptian (EGY), Levantine (LEV), Arabizi, and noisy Arabic. These are not repeated AraSafe base estimates; for example, Fanar's MSA 5.8/93.3 is measured on this boundary distribution.}
\label{tab:appendix-boundary-set}
\begin{tabular}{llrr}
\toprule
Model & Variant & B & H \\
\midrule
ALLaM-7B   & MSA     & 10.8 & 94.8 \\
           & EGY     & 31.3 & 95.2 \\
           & LEV     & 17.1 & 94.8 \\
           & Arabizi & 45.0 & 70.0 \\
           & Noisy   & 33.8 & 91.9 \\
\midrule
Fanar-9B   & MSA     &  5.8 & 93.3 \\
           & EGY     &  7.7 & 88.1 \\
           & LEV     &  5.2 & 88.1 \\
           & Arabizi &  3.3 & 43.8 \\
           & Noisy   &  8.7 & 86.2 \\
\midrule
Llama-3-8B & MSA     & 16.0 & 94.3 \\
           & EGY     & 24.8 & 86.7 \\
           & LEV     & 22.1 & 88.6 \\
           & Arabizi & 49.6 & 83.8 \\
           & Noisy   & 30.8 & 89.0 \\
\midrule
Qwen2.5-3B & MSA     &  6.0 & 86.2 \\
           & EGY     &  8.3 & 71.4 \\
           & LEV     &  5.8 & 78.1 \\
           & Arabizi & 14.6 & 22.9 \\
           & Noisy   & 14.0 & 78.6 \\
\midrule
Qwen2.5-7B & MSA     &  4.0 & 88.6 \\
           & EGY     &  7.5 & 73.8 \\
           & LEV     &  5.6 & 77.1 \\
           & Arabizi &  8.3 & 30.0 \\
           & Noisy   &  9.8 & 72.9 \\
\bottomrule
\end{tabular}
\end{minipage}\end{center}

\Needspace{18\baselineskip}
\section{Additional Diagnostic Figures}
\label{app:diagnostic-figures}

This appendix gives the full-resolution figures summarized in the main text, grouped by topic.

\subsection{Base and Selected Operating Points}

\begin{figure}[H]
  \centering
  \includegraphics[width=\linewidth]{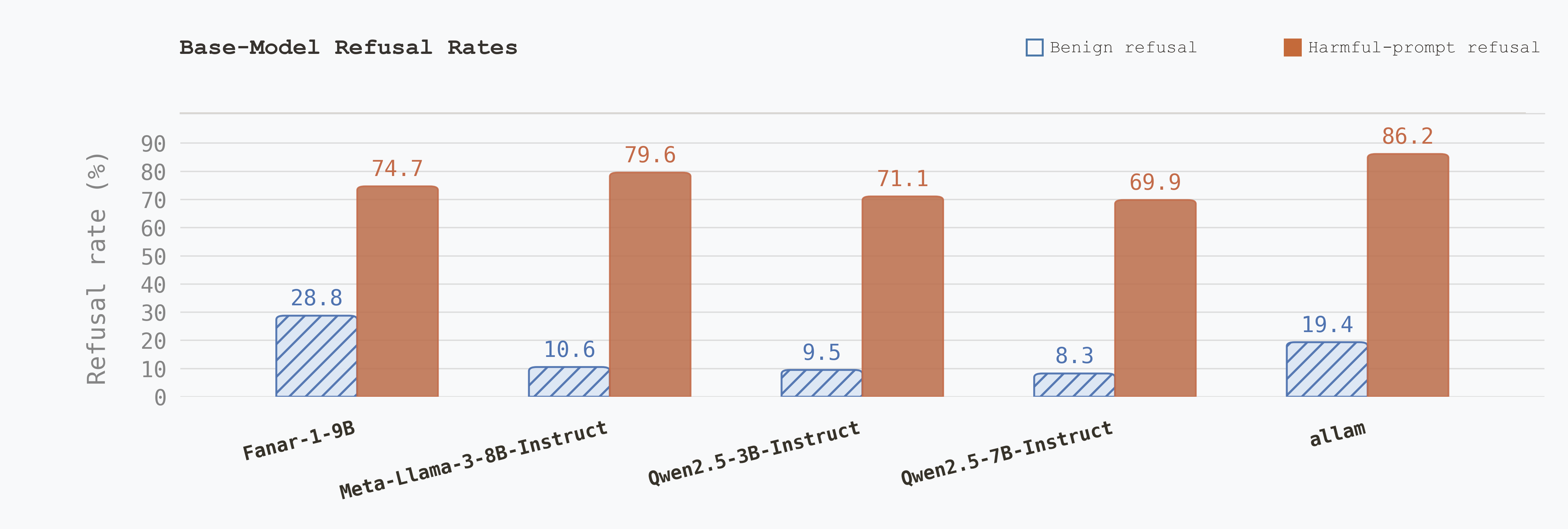}
  \caption{Base-model AraSafe refusal rates. Hatched blue bars show benign refusal $\safe$, and orange bars show harmful-prompt refusal $\unsafe$. Values above the bars are percentages.}
  \label{fig:base-bars}
\end{figure}

\begin{figure}[H]
  \centering
  \includegraphics[width=\linewidth]{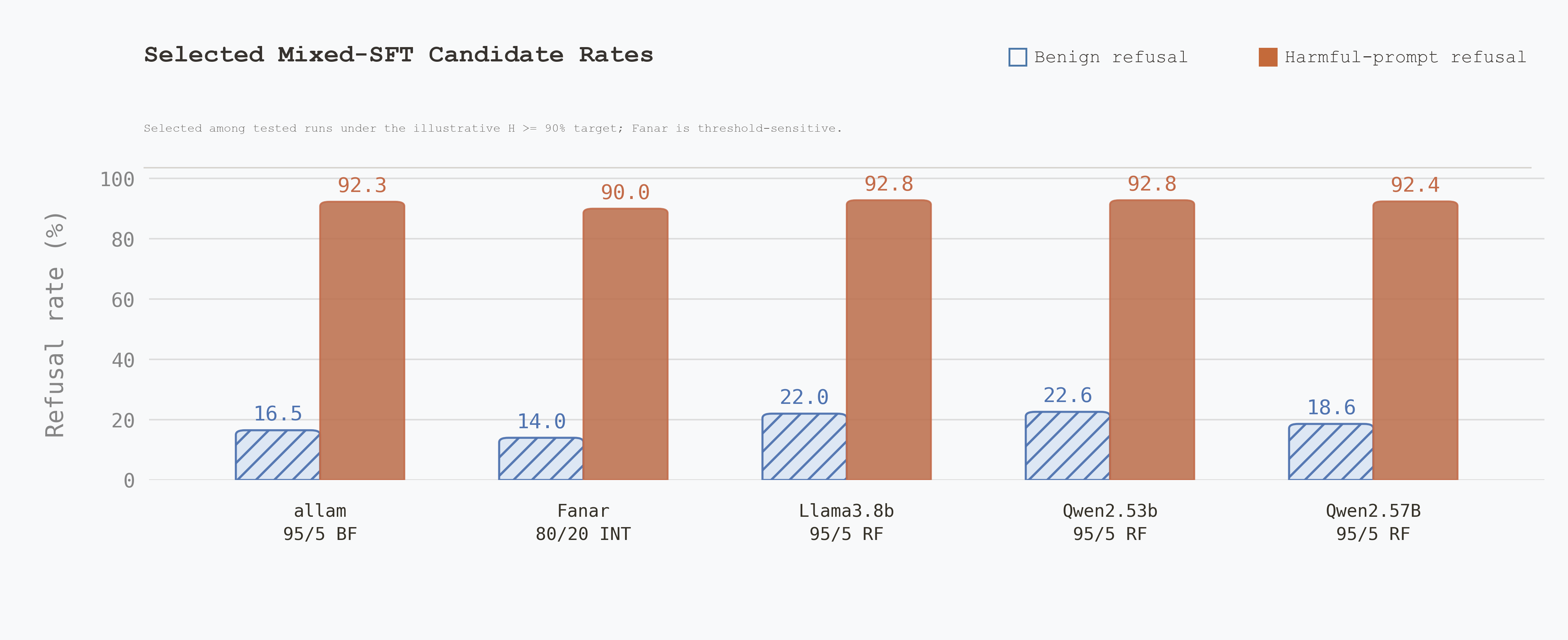}
  \caption{Selected mixed-SFT operating points. Hatched blue bars show benign refusal $\safe$, and orange bars show harmful-prompt refusal $\unsafe$. Ordering labels identify runs only; the ordering comparison is null. Fanar is threshold-sensitive across seeds.}
  \label{fig:selected}
\end{figure}

\subsection{Training and Calibration Detail}

\begin{figure}[H]
  \centering
  \includegraphics[width=\linewidth]{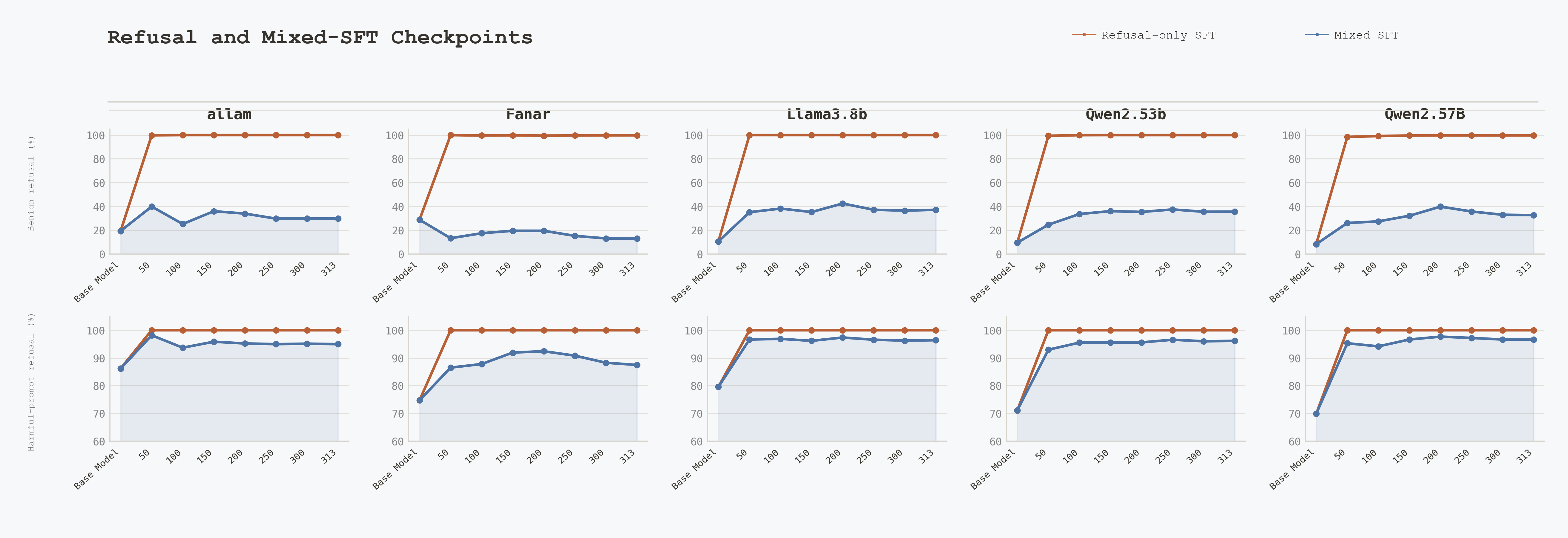}
  \caption{Refusal-only and mixed-SFT trajectories across checkpoints. The top row reports benign refusal $\safe$, and the bottom row reports harmful-prompt refusal $\unsafe$. Lines connect measured checkpoints only.}
  \label{fig:checkpoints}
\end{figure}

\subsection{Internal Diagnostics}

\begin{figure}[H]
  \centering
  \includegraphics[width=\linewidth]{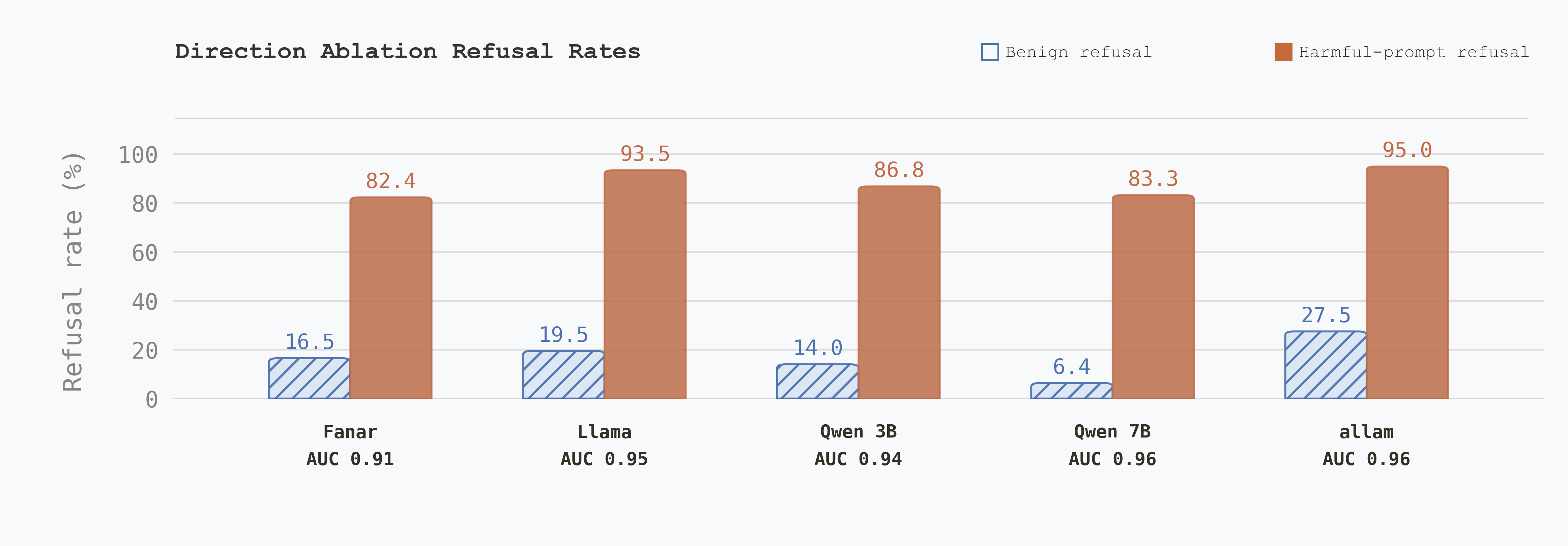}
  \caption{Operating points induced by calibrated refusal-direction thresholds. Hatched blue bars show benign refusal $\safe$, orange bars show harmful-prompt refusal $\unsafe$, and direction AUC appears below each model.}
  \label{fig:directions}
\end{figure}

\begin{figure}[H]
  \centering
  \includegraphics[width=\linewidth]{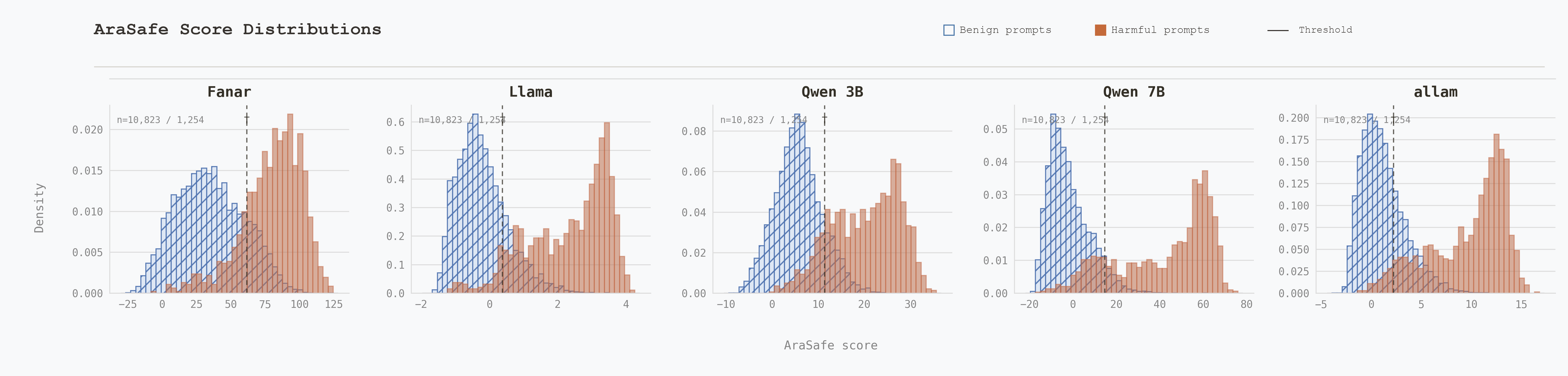}
  \caption{Per-model AraSafe refusal-direction score distributions for 10{,}823 benign and 1{,}254 harmful prompts. Histograms show densities for both groups, and dashed lines mark calibrated thresholds. Score scales are model-specific.}
  \label{fig:score-dist}
\end{figure}

\Needspace{10\baselineskip}
\section{Qualitative Direction Examples}
\label{app:examples}

Harmful prompts and enabling model continuations are redacted. Qualitative evidence is reported only as labels indicating whether the original refusal was preserved or harmful ablated content was removed.

\begin{table}[H]
  \centering
  \small
  \caption{Redacted qualitative direction ablation examples. Harmful prompts and enabling ablated content are omitted; benign rows are retained only when they do not expose harmful content.}
  \label{tab:examples}
  \begin{tabular}{lll}
    \toprule
    Row type & Prompt & Observed behavior \\
    \midrule
    Benign & Sensitive prompt preserved & Refusal boundary remains model-specific \\
    Benign & Clean prompt preserved & Access is retained only at some operating points \\
    Harmful & Prompt redacted & Original refusal preserved; enabling content redacted \\
    Harmful & Prompt redacted & Original refusal preserved; enabling content redacted \\
    \bottomrule
  \end{tabular}
\end{table}

\section{Base Model IDs and Release Plan}
\label{app:ids}

The base model IDs are \nolinkurl{Qwen/Qwen2.5-3B-Instruct}, \nolinkurl{Qwen/Qwen2.5-7B-Instruct}, \nolinkurl{meta-llama/Meta-Llama-3-8B-Instruct}, \nolinkurl{QCRI/Fanar-1-9B}, and \nolinkurl{ALLaM-AI/ALLaM-7B-Instruct-preview}.
The ALLaM Hugging Face page redirects to \nolinkurl{humain-ai/ALLaM-7B-Instruct-preview}. The planned release includes training and evaluation code, configurations, seeds, judge prompts, anonymized audit labels, transformation or reconstruction scripts, and permitted adapter or checkpoint identifiers. No public artifact URL is available in the current record, so the paper does not claim that these materials are already released.

\end{document}